\documentclass[runningheads]{llncs}
\usepackage{graphicx}

\usepackage{tikz}
\usepackage{comment}
\usepackage{amsmath,amssymb} 
\usepackage{color}
\usepackage{booktabs}
\usepackage[accsupp]{axessibility}  

\usepackage[left=22mm,top=22mm]{geometry}

\begin{document}
\pagestyle{headings}
\mainmatter
\def\ECCVSubNumber{1139}  

\title{FaceSigns: Semi-Fragile Neural Watermarks for Media Authentication and Countering Deepfakes} 

\titlerunning{FaceSigns}
%
\author{\text{*}Paarth Neekhara$^1$, \text{*}Shehzeen Hussain$^1$, Xinqiao Zhang$^{1,2}$, Ke Huang$^2$, \\ Julian McAuley$^1$, Farinaz Koushanfar$^1$\\
{\tt\footnotesize \{pneekhar,ssh028\}@eng.ucsd.edu} \\
\text{*} Equal contribution\\
}
\authorrunning{Neekhara and Hussain et al.}
%
\institute{University of California San Diego \and
San Diego State University
}
\maketitle

\begin{abstract}
Deepfakes and manipulated media are becoming a prominent threat due to the recent advances in realistic image and video synthesis techniques. There have been several attempts at combating Deepfakes using machine learning classifiers. However, such classifiers do not generalize well to black-box image synthesis techniques and 
have been shown
to be vulnerable to adversarial examples. To address these challenges, we introduce a deep learning based semi-fragile watermarking technique that allows media authentication by verifying an invisible secret message embedded in the image pixels. 
Instead of identifying and detecting fake media using visual artifacts, we propose to proactively embed a semi-fragile watermark into a real image so that we can prove its authenticity when needed. Our watermarking framework is designed to be fragile to facial manipulations or tampering while being robust to 
benign image-processing operations such as
image compression, scaling, saturation, contrast adjustments etc. This allows 
images shared over the internet 
to retain the verifiable watermark as long as face-swapping or any other Deepfake modification technique is not applied. We demonstrate that FaceSigns can embed a 128 bit secret as an imperceptible image watermark that can be recovered with a high bit recovery accuracy at several compression levels, while being non-recoverable when unseen Deepfake manipulations are applied. 
For a set of unseen benign and Deepfake manipulations studied in our work, FaceSigns can reliably detect manipulated content with an AUC score of $0.996$ which is significantly higher than prior image watermarking and steganography techniques\footnote{Inference Demo: \url{https://github.com/paarthneekhara/FaceSignsDemo}}.

\end{abstract}

\section{Introduction}

Media authentication, 
despite having been a
long-term challenge, has become even more difficult with the advent of deep learning based generative models.  
Deep Neural Network (DNN) based media synthesis techniques~\cite{thies2016face2face,DeepFakesgit,faceswap,zakharov2019few,nirkin2019fsgan,karras2019style}, have enabled the creation of high-quality convincing fake photos and videos. One such example of synthetic videos is 
Deepfakes, in which a subject's face is swapped with a target face to create convincingly realistic footage of events that never occurred~\cite{faceforensicsiccv,advdeepfakes}. 
Such manipulated videos can fuel misinformation, defame individuals and reduce trust in media~\cite{dfsurvey}. 
Recent Deepfake detection methods rely on DNN based classifiers to distinguish synthetic videos from real videos~\cite{faceforensicsiccv,dolhansky2020deepfake}. However, with advances in Deepfake synthesis methods, DNN based classification methods are failing to stay a step ahead in the race to reliably detect Deepfakes. This is largely because classifiers trained in a supervised manner on existing Deepfake generation methods cannot be reliably secure against black-box Deepfake generation methods. Moreover, the current best performing Deepfake detectors can be easily bypassed by attackers using adversarial examples~\cite{advdeepfakes,gandhi2020adversarial,neekhara2020adversarial}.

\begin{figure}[h]
    \centering
    \includegraphics[width=0.8\linewidth]{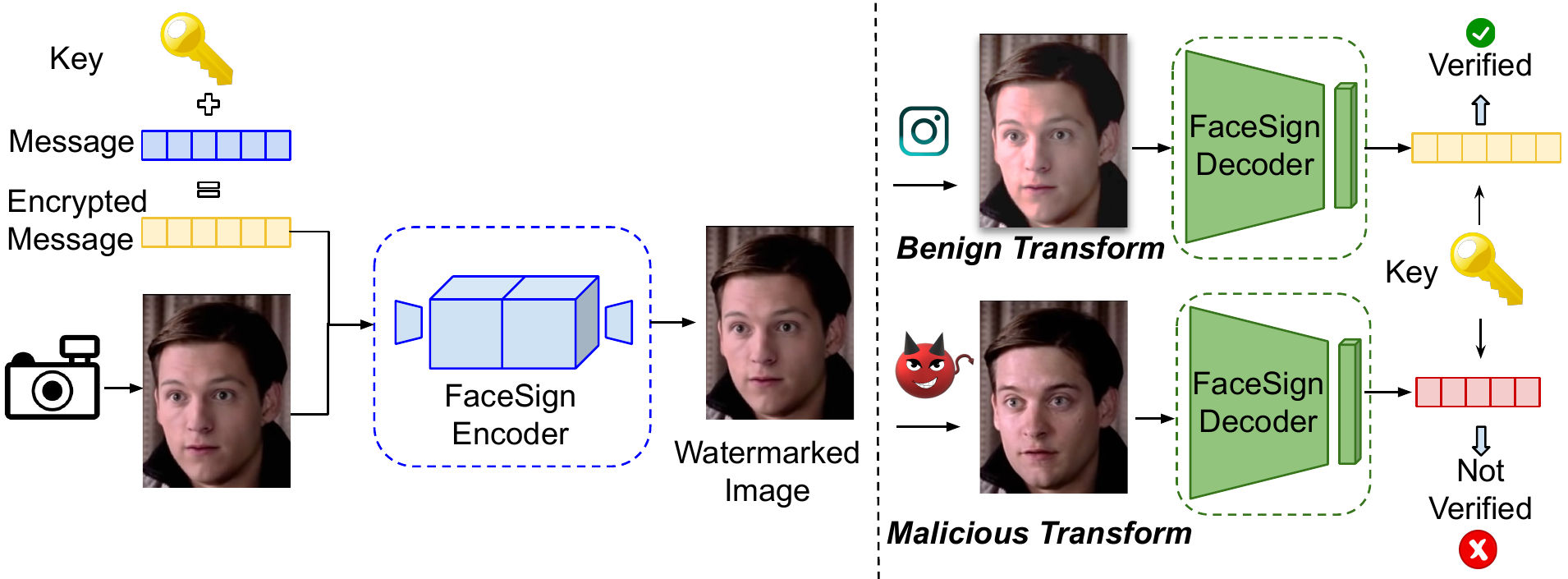}
    \caption{Overview of \textit{FaceSigns} Watermarking framework: The encoder network embeds a secret encrypted message into a given image as an imperceptible watermark that is designed to be robust against benign image transformations and photo editing tools but fragile towards malicious image manipulations such as Deepfakes.
    }
    \label{fig:introdiag}
\end{figure}

We believe that proactively embedding a secret verifiable message into the images when they are captured from a device, can establish the provenance of real images and videos and address the limitations of Deepfake classifiers.
Several prior works have explored digital image watermarking and deep learning based steganography techniques~\cite{cox2002digital,fridrich2009steganography,tancik2020stegastamp,zhu2018hidden,luo2020distortion} to hide secret messages in image pixels. However, these works are either fragile to basic image processing operations such as compression and color adjustments or overly robust to the point that the secret can be recovered even after occluding major portions of the embedded image~\cite{tancik2020stegastamp}. In fact, we experimentally demonstrate that past works on robust neural watermarks~\cite{tancik2020stegastamp,zhu2018hidden} can recover messages even from images that have undergone Deepfake manipulations. 

To address the above challenges of Deepfake classifiers and neural watermarking frameworks, we introduce~\textit{FaceSigns} --- a deep learning based semi-fragile watermarking system that embeds a recoverable message as an imperceptible perturbation in the image pixels. 
The watermark can contain a secret message or device-specific codes which can be used for authenticating images and videos. The desirable property of the watermark is that it should break if a malicious facial manipulation such as a Deepfake transformation is applied to the image, but it should be robust against harmless image transformations such as image compression, color and lighting adjustments which are commonly applied on pictures before uploading them to online sharing platforms. To achieve this goal, we develop an encoder-decoder based training framework that encourages message recovery under benign transformations and discourages message recovery if the watermark is tampered in the facial regions of the image. In contrast to hand-designed pipelines used in previous work for semi-fragile watermarking~\cite{lin2000detection,ho2004semi,yang2009semi,preda2015watermarking,bhalerao2021secure}, our framework is end-to-end and learns to be robust to a wide range of real-world digital image processing operations such as social media filters and compression techniques, while being fragile to various Deepfake tampering techniques. The technical contributions of our work are as follows:
\begin{enumerate}
    \item We develop a neural semi-fragile image watermarking framework as a proactive defense against Deepfakes. To the best of our knowledge, our work is the first deep learning based semi-fragile watermarking system that can certify the authenticity of digital media.
    \item We design a differentiable procedure to simulate facial watermark tampering during training such that our framework can achieve selective fragility against unseen malicious transformations (Section~\ref{sec:malicious}).
    \item For a set of previously unseen benign and malicious image transformations, FaceSigns achieves the goal of selective fragility and reliably detects Deepfake manipulations with an AUC score of $\mathbf{0.996}$ which is significantly higher than alternate robust and semi-fragile watermarking frameworks.
\end{enumerate}

\section{Background}


\subsection{Facial Forgery}
Until recently, the ease of generating manipulated photos and videos has been limited by manual editing tools. However, since the advent of deep learning there has been significant work in developing new techniques for automatic digital forgery. 
DNN based facial manipulation methods~\cite{faceswap,simswap,DeepFakesgit,choi2020starganv2} 
operate end-to-end on a source video and target face, and require minimal human expertise to generate fake videos in real-time. 
In our work we show effectiveness against popular GAN based Deeepfake generation methods - SimSwap~\cite{simswap} and FSFT~\cite{fsft}, and a classical computer graphics based face replacement approach - FaceSwap~\cite{faceswap}. 

The best performing Deepfake detectors~\cite{afchar2018mesonet,dolhansky2019deepfake,faceforensicsiccv,selim,wang2020cnn} rely on  convolutional neural network (CNN) based architectures. 
Such Deepfake detectors model Deepfake detection as a per-frame binary classification problem, using a face-tracking method prior to CNN classification to effectively detect facial forgeries in both uncompressed and compressed videos. While CNN based classifiers achieve promising detection accuracy on a fixed in-domain test set of real and fake videos, they suffer from two main drawbacks: 1) Lack of generalizability to unseen Deepfake synthesis techniques. 2) Vulberability to adversarial examples in both black-box and white-box attack settings. We refer the reader to past works~\cite{gandhi2020adversarial,ACM_adv_deepfakes,advdeepfakes,9105991} that explore such limitations of CNN Deepfake detectors.


\subsection{Digital Watermarking}
Digital watermarking~\cite{cox2002digital} similar to steganography~\cite{fridrich2009steganography}, is the task of embedding information into an image in a visually imperceptible manner. 
These techniques broadly seek to generate three different types of watermarks: fragile~\cite{di2019fragile,bhalerao2021secure}, robust~\cite{cox1997secure,pereira1999template,bi2007robust,shehab2018secure,zhu2018hidden,pereira2000robust} and semi-fragile~\cite{lin2000detection,sun2002svd,yu2017review}
watermarks.
Fragile and semi-fragile watermarks are primarily used to certify the integrity and authenticity of image data.
Fragile watermarks are used to achieve accurate authentication of digital media, where even a one-bit change to an image will lead it to fail the certification system.
In contrast, \textit{robust} watermarks aim to be recoverable under several image manipulations, in order to allow media producers to assert ownership over their content even if the video is redistributed and modified.
Semi-fragile watermarks combine the advantages of both robust and fragile watermarks, and are mainly used for fuzzy authentication of digital images and identification of image tampering~\cite{yu2017review}. 
The use of semi-fragile watermarks is justified by the fact that images and videos are generally transmitted and stored in a compressed form, which should not break the watermark. However when the image gets tampered, the watermark should also get damaged, indicating image tampering.

Several past works have proposed hand-engineered pipelines to embed semi-fragile watermark information in the spatial and frequency (transform) domain of images and videos. In the spatial domain, the pixels of digital images are processed directly using block-based embedding~\cite{bhalerao2021secure} and least significant bits modification~\cite{xiao2008semi,yang2009semi} to embed watermarks. In the frequency
domain, the watermark can be embedded by modifying the coefficients produced with transformations such as the Discrete Cosine Transform (DCT)~\cite{preda2015watermarking,ho2004semi} and Discrete Wavelet Transform~\cite{li2015semi,benrhouma2015tamper,shefali2007information}. 
However, we demonstrate in our experiments the major limitations of traditional approaches lies in higher visibility of the embedded watermarks, increased distortions in generated images and low robustness to compression techniques like JPEG transforms. Moreover, these works have not been designed to be fragile against Deepfake manipulations.

More recently, CNNs have been used to provide an end-to-end solution to the watermarking problem. They replace hand-crafted hiding procedures with neural network encoding~\cite{baluja2017hiding,hayes2017generating,zhu2018hidden,zhang2019invisible,luo2020distortion,tancik2020stegastamp}. Notably, both StegaStamp~\cite{tancik2020stegastamp} and HiDDeN~\cite{zhu2018hidden} propose frameworks to embed robust watermarks that can hide
and transmit data in a way that is robust to various real-world transformations.
All of these works focus on generating robust watermarks, with the goal of ensuring robustness and recovery of the embedded secret information under various physical and digital image distortions. We empirically demonstrate that these techniques are unable to generate semi-fragile watermarks and are therefore not suitable for identifying tampered media such as Deepfakes.

\section{Methodology}

\begin{figure*}
    \centering
    \includegraphics[width=0.9\linewidth]{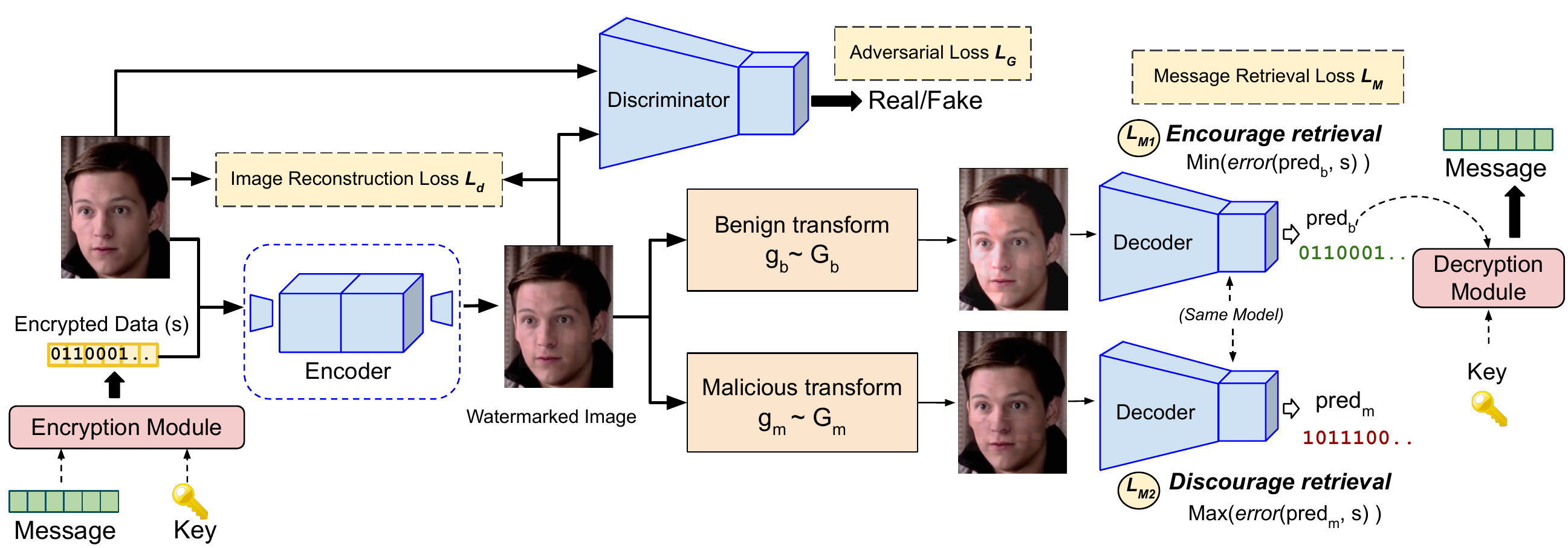}
    
    \caption{Model overview: The encoder and decoder networks are trained by encouraging message retrieval from watermarked images that have undergone benign transformations and discouraging retrieval from maliciously transformed watermarked images. Image reconstruction and adversarial loss from the discriminator ensure the imperceptibility of the watermark.
    }
    \label{fig:model_diagram}
\end{figure*}

We aim to develop an image watermarking framework that is robust to a set of benign image transformations (eg. brightness and contrast adjustments, resizing, JPEG compression (etc.)) while being fragile to facial manipulations like GAN-based face swapping or conventional face editing. Additionally, it is desirable to have an imperceptible watermark so that devices can store only the watermarked images without revealing the original image to the end user. To this end, our system consists of three main components: an encoder network $E_\alpha$, a decoder network $D_\beta$ and an adversarial discriminator network $A_\gamma$ where $\alpha, \beta \text{ and } \gamma$ are learnable parameters.

The encoder network $E$ takes as input an image $x$ and a bit string $s \in \{0, 1\}^L$ of length $L$, and produces an encoded (watermarked) image $x_w$. That is, $x_w=E(x, s)$. The watermarked image then goes through two image transformation functions --- one sampled from a set of benign transformations ($g_b \sim G_b$) and the other sampled from a set of malicious transformations  ($g_m \sim G_m$) to produce a benign image $x_b=g_b(x_w)$ and a malicious image $x_m=g_m(x_w)$. The benign and malicious watermarked images are then fed to the decoder network which predicts the messages $s_b=D(x_b)$ and $s_m=D(x_m)$ respectively.

For optimizing secret retrieval during training, we use the $L_1$ distortion between the predicted and ground-truth bit strings.
The decoder is encouraged to be robust to benign transformations by minimizing the message distortion $L_1(s, s_b)$; and fragile for malicious manipulations by maximizing the error $L_1(s, s_m)$. Therefore the secret retrieval error for an image $L_M(x)$ is obtained as follows:
\begin{equation}
L_M(x) = L_1(s, s_b) - L_1(s, s_m)
\end{equation}

The watermarked image is encouraged to look visually similar to the original image by optimizing three image distortion metrics: $L_1$, $L_2$ and $L_\textit{pips}$~\cite{zhang2018unreasonable} distortions. Additionally, we use an adversarial loss $L_G(x_w) = \log(1 - A(x_w))$ from the discriminator which is trained simultaneously to distinguish original images from watermarked images. That is, our image reconstruction loss $L_\textit{img}$ is obtained as follows:
\begin{equation}
\begin{split}
& L_\textit{d}(x, x_w) = L_1(x, x_w) + L_2(x, x_w) + c_p L_\textit{pips}(x, x_w) \\
& L_\textit{img}(x, x_w) = L_\textit{d}(x, x_w) + c_g L_G(x_w)
\end{split}
\end{equation}

Therefore, the parameters $\alpha,\beta$ of the encoder and decoder network are trained using mini-batch gradient descent to optimize the following loss over a distribution of input messages and images: 
\begin{equation}
\mathbb{E}_{x, s, g_b, g_m} [ L_\textit{img}(x, x_w) + c_M L_M(x)  ]
\end{equation}

The discriminator parameters $\gamma$ are trained to distinguish original images $x$ from watermarked images $x_w$ as follows:
\begin{equation}
\mathbb{E}_{x, s} [\log(1 - A(x)) + \log(A(x_w)) ]
\end{equation}

In the above equations, $c_p$, $c_g$, $c_M$ are scalar coefficients for the respective loss terms. We refer the readers to our supplementary material for the values we use for these coefficients and other implementation details.

\subsubsection{Message encoding} The encoder network accepts watermarking data as a bit string $s$ of length $L$. This watermarking data can contain information about the device that captured the image or a secret message that can be used to authenticate the image.
To prevent adversaries (who have gained white-box access to the encoder network) from encoding a target message, we can encrypt the message using symmetric or asymmetric encryption algorithms or hashing. In our experiments, we embed encrypted messages of size 128 bits which allows the network to encode $2^{128}$ unique messages. We discuss the possible threats and defenses to our watermarking framework in Section~\ref{sec:threatmodels}.

\subsection{Network Architectures}
\label{sec:networkarch}
Our encoder and decoder networks are based on the U-Net CNN architecture~\cite{ronneberger2015u,isola2017image,tancik2020stegastamp} and operate on $256\times256$ images. 
The encrypted message $s$, which is an $L$ length bit string, is first projected to a tensor $s_{\textit{Proj}}$ of size $96\times96$ using a trainable fully connected layer; then resized to $256 \times 256$ using bilinear interpolation and finally added as the fourth channel to the original RGB image to be fed as an input to the encoder network. The encoder U-Net contains 8 downsampling and 8 unsampling layers. We modify the original U-Net architecture and replace the transposed convolution in the upsampling layers with convolutions followed by nearest-neighbour upsampling as per the recommendations given by~\cite{odena2016deconvolution}. In our preliminary experiments, we found this change to significantly improve the image quality and training speed of our framework. 
The decoder network also follows the U-Net architecture similar to our encoder network. The decoder U-Net first outputs a $256\times256$ intermediate output, which is downsized to $96\times96$ using bilinear down-sampling to produce $s_\textit{ProjDecoded}$ 
and then projected to a vector of size $L$ using a fully connected layer followed by a sigmoid layer to scale values between $0$ and $1$. 

For the discriminator network, we use the patch discriminator from~\cite{isola2017image}. 
The discriminator is trained to classify each $N\times N$ image patch as real or fake. We average discriminator responses across all patches to obtain the discriminator output. Our discriminator network consists of three convolutional blocks of stride 2 thereby classifying patches of size $32\times 32$.

\subsection{Transformation functions}
The choice of benign and malicious transformation functions is critical to achieve selective fragility and robustness of the watermark. While we can only use a limited set of image transformations during training, the list of possible benign and malicious transforms in real-world setting is non exhaustive. 
In our experiments (Section~\ref{sec:robustnessfragility}), we demonstrate that by incorporating the below described transformation functions, we are able to generalize to unseen benign and malicious transformations. 

\subsubsection{Benign Transforms}
Our goal is to authenticate real images shared over online platforms that generally undergo diverse color or lighting adjustments (e.g.~Instagram filters). Therefore, to approximate standard image processing distortions, we apply a diverse set of differentiable benign image transformations ($G_b$) to our watermarked images during training:

 \noindent \textbf{1. Gaussian Blur:} We convolve the original image with a Gaussian kernel $\mathit{k}$. This transform is given by $t(x)= k \ast x $ where $\ast$ is the convolution operator. We use kernel sizes ranging from $\mathit{k} = 3$ to $\mathit{k} = 7$
    
 \noindent \textbf{2. JPEG compression:} Digital images are usually stored in a lossy format such as JPEG. We approximate JPEG compression with the differentiable JPEG function proposed in~\cite{shin2017jpeg}. During training, we apply JPEG compression with quality $40$, $60$ and $80$.
    
 \noindent \textbf{3. Saturation adjustments:} To account for various color adjustments from social media filters, we randomly linearly interpolate between the original (full RGB) image and its grayscale equivalent.
    
 \noindent \textbf{4. Contrast adjustments:} We linearly rescale the image histogram using a contrast factor $\sim \mathcal{U}[0.5, 1.5]$
    
 \noindent \textbf{5. Downsizing and Upsizing:} The image is first downsized by a factor $\textit{scale}$ and then up-sampled by the same factor using bilinear upsampling. We use $\textit{scale} \sim \mathcal{U}[2, 5]$
    
 \noindent \textbf{6. Translation and rotation:} The image is shifted horizontally and vertically by $n_h$ and $n_w$ pixels where $n_h, n_w \sim \mathcal{U}[-10, 10]$  and rotated by $r$ degrees where $r \sim \mathcal{U}[-10, 10]$.
 
For each mini-batch iteration, we sample one transformation function from the above list (and an Identity transform) and apply it to all the images in the batch. 

\subsubsection{Malicious Transforms}
\label{sec:malicious}
\begin{figure}[h]
    \centering
    \includegraphics[width=1.0\linewidth]{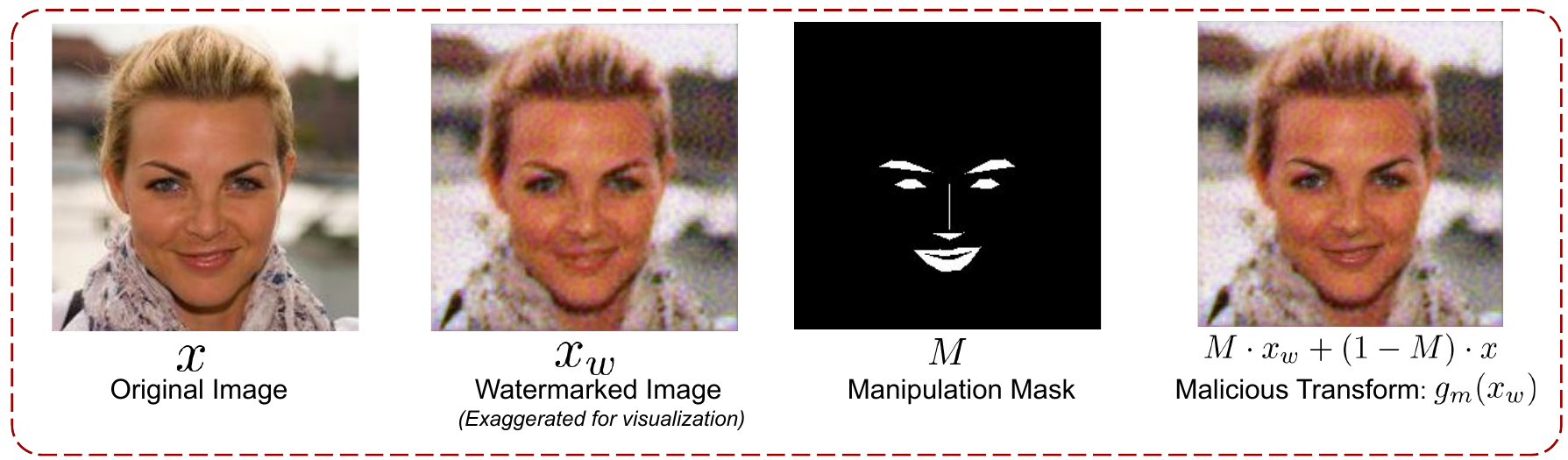}
    
    \caption{Malicious Transform: To simulate facial tampering during training, the watermark is partially removed from the facial features of the image such as eyes, nose and lips. 
    }
    \label{fig:manipulation}
    
\end{figure}
Our semi-fragile watermarks have to be unrecoverable when facial tampering such as face swapping or GAN based manipulations are applied. 
Explicitly including Deepfake manipulation systems in our training pipeline can create same drawbacks as supervised Deepfake classifiers and lead to lack of generalization to unseen Deepfake synthesis techniques.

The common operation across Deepfake techniques is modifying facial features to achieve the appearance of the target identity. 
Therefore, we model malicious tampering as a transformation function $g_m$ that selectively modifies the watermark in certain regions of the face.
For each image, we initialize a mask $M_{h\times w\times c}$ of all ones. Next, we extract the facial feature polygons for \textit{eyes}, \textit{nose} and \textit{lips} and set the values for all pixels inside the polygons to a small watermark retention percentage $w_r \in [0,1]$. 
That is, $M[i,j,:] = w_r$ for all $i,j$ in the facial feature polygons. Finally, the maliciously transformed image $g_m(x_w)$ is obtained as follows:

$$ g_m(x_w) = M\cdot x_w + (1-M)\cdot x $$

Figure~\ref{fig:manipulation} illustrates the malicious transform procedure. 
The underlying assumption is that Deepfake manipulations modify facial features thereby tampering the watermark in those regions. 
\section{Experiments}
\label{sec:experiments}

\subsection{Datasets and Experimental Setup}
We conduct our experiments on the CelebA dataset~\cite{liu2015faceattributes} which is a large scale database of over $200,000$ face images of $10,000$ unique celebrities. We set aside $1000$ images for testing the watermarking models and split the remaining data into $80\%$ training and $20\%$ validation. We train our models for $100K$ mini-batch iterations with a batch-size of $64$ and use an Adam optimizer with a fixed learning rate of $2e-4$. 
All our models are trained using images of size $256\times256$ which are obtained after center-cropping and resizing the CelebA images.
We conduct experiments with message length $L=128$.
To evaluate the effectiveness of using transformation functions during training, we conduct an ablation study by training a \textit{FaceSigns (No Transform)} model that does not incorporate any input transformations and a \textit{FaceSigns (Robust)} model that uses only benign transformations during training.
We evaluate watermarking techniques primarily on the following aspects:

\noindent\textbf{1. Imperceptibility:}  We compare the original and watermarked images to compute: \textbf{peak signal to noise ratio (PSNR)} and \textbf{structural similarity index (SSIM)}. Higher values for both PSNR and SSIM are desirable for a more imperceptible watermark.

\noindent \textbf{2. Robustness and Fragility:} To measure the robustness and fragility of the watermarking system we measure the \textbf{bit recovery accuracy (BRA)} of the bit string $s$ when unseen (not used in training) benign and malicious image transformations are applied. 
For robustness, it is desirable to have a high BRA against benign transformations like social media filters and image compression. 
For fragility against Deepfake synthesis techniques, it is desirable to have a low BRA when a Deepfake facial manipulation is applied. 
To make a fair comparison with past works, we do not apply any bit error correcting codes while calculating the BRA and compare the input string $s$ with the raw decoder output.
A detector can classify an input as manipulated if the BRA of the decoded message is below a set threshold and benign if the BRA is more than the threshold. We measure the performance of such a detector using the \textbf{AUC score - Area under the ROC curve}.
    
\noindent \textbf{3. Capacity:} Capacity measures the amount of information that can be embedded in the image. We measure the capacity as the \textbf{bits per pixel (BPP)} that is the number of bits of the encrypted message embedded per pixel of the image which is simply $=L/(\mathit{HWC})$.

It is important to note the trade-off between the above metrics---e.g. models with higher capacity, sacrifice on the imperceptibility or bit recovery accuracy. Similarly, more robust models sacrifice capacity or imperceptibility. We compare our watermarking framework against three prior works on image watermarking --- a DCT based semi-fragile watermarking system~\cite{ho2004semi} and two neural image watermarking systems HiDDeN~\cite{zhu2018hidden} and StegaStamp~\cite{tancik2020stegastamp}. 
Both HiDDeN and StegaStamp embed a bit string message into a square RGB image while ensuring robustness to a set of image transformations.   



\setlength\tabcolsep{6pt}
\begin{table}[h]
\centering
\begin{tabular}{@{}l|rrr|cc@{}}
\multicolumn{1}{r}{} & \multicolumn{3}{c|}{\emph{Capacity}} & \multicolumn{2}{c}{\emph{Imperceptibility}}
\\

\midrule
Method & H,W & $L$ & $\text{BPP}$ & PSNR & SSIM  \\ 
\midrule

SemiFragile DCT~\cite{ho2004semi} & $128$ & $256$ & $5.2\text{e-}3$ & $22.49$ & $0.871$ \\
HiDDeN~\cite{zhu2018hidden} & $128$ & $30$ & $6.1\text{e-}4$ & $27.57$ & $0.934$ \\
StegaStamp~\cite{tancik2020stegastamp} & $400$ & $100$ & $2.0\text{e-}4$  & $29.39$ & $0.925$\\\midrule
FaceSigns (No Transform) & $256$ & $128$ & $6.5\text{e-}4$ & $36.38$ & $0.973$ \\
FaceSigns (Robust) & $256$ & $128$ & $6.5\text{e-}4$ & $35.91$ & $0.971$\\
FaceSigns (Semi-Fragile) & $256$ & $128$ & $6.5\text{e-}4$ & $36.08$ & $0.975$\\

\bottomrule
\multicolumn{1}{c}{}
\end{tabular}
\caption{Capacity and imperceptibility metrics of different watermarking systems. $H,W$ indicate the height and width of the input image.}
\label{tab:basicompare}
\end{table}


\begin{figure*}
    \centering
    \includegraphics[width=1.0\linewidth]{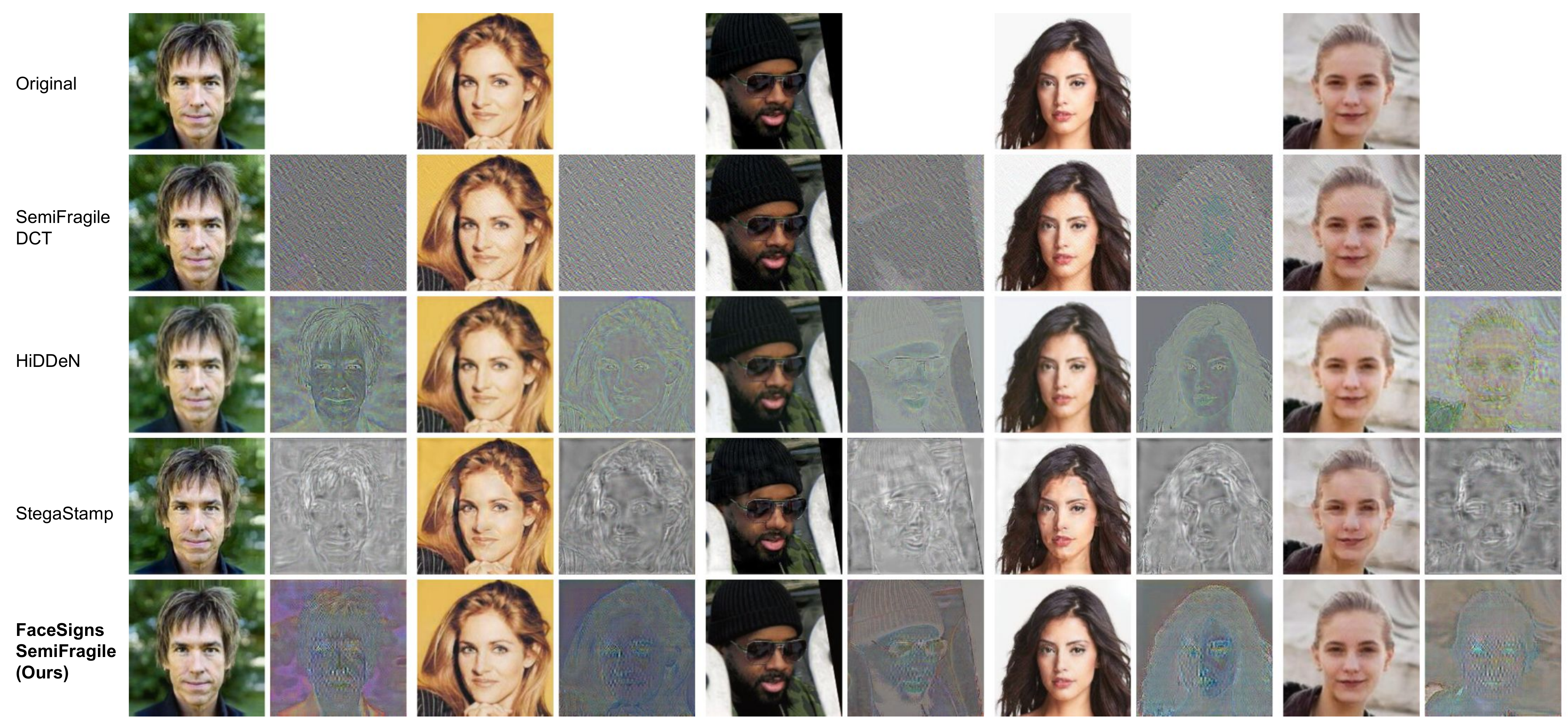}
    \caption{Examples of original and watermarked images using prior works and our \textit{FaceSigns (Semi-Fragile)} model. The image perturbation has been linearly scaled between 0 and 1 for visualization. 
    The quantitative metrics evaluating the capacity and imperceptibility of the watermark are reported in Table~\ref{tab:basicompare}.
    }
    \label{fig:comparisonresults}
\end{figure*}
\subsection{Imperceptibility and Capacity}
We report the image similarity and capacity metrics of different watermarking techniques in Table~\ref{tab:basicompare}. 
We find that even at a higher message capacity,  FaceSigns can encode messages with better imperceptibility as compared to StegaStamp and HiDDeN. As noted by the authors of StegaStamp and visible in our Figure~\ref{fig:comparisonresults}, the residual added by their model is perceptible in large low frequency regions of the image.
We believe that this is primarily due to the difference in our network architecture choices. In our initial experiments, we found that using a UNet architecture for the decoder with an intermediate message reconstruction loss described in Section~\ref{sec:networkarch}, performed significantly better than a down-sampling CNN architecture used in prior work.
Additionally, we use nearest neighbour upsampling instead of transposed convolutions in our U-Net architectures which helps reduce the perceptibility of the watermark by removing upsampling artifacts.

\subsection{Robustness and Fragility}
\label{sec:robustnessfragility}

To study the robustness and fragility of different DNN based watermarking techniques, we transform the watermarked images using unseen benign and malicious transformations and then attempt to decode the message from the transformed message. In this evaluation we consider out-of-domain and practical transformations which are generally applied before images are uploaded to the internet. 
\begin{figure}
    \centering
    \includegraphics[width=1.0\linewidth]{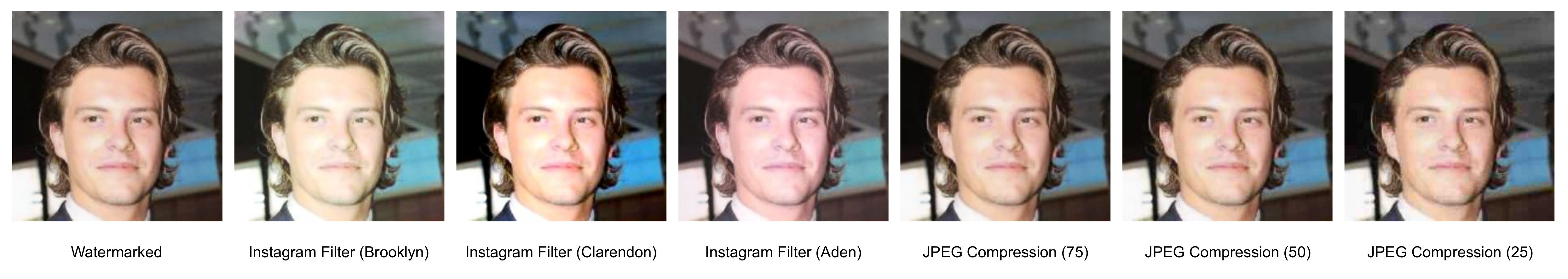}
    
    \caption{ Watermarked images with unseen benign transformations applied. Benign transformations depicted in this diagram include Instagram filters~\cite{pilgramfilter} Brooklyn, Clarendon, Aden and various levels of JPEG compression
    }
    \label{fig:filters}
\end{figure}

For benign transformations, we compress the image using different levels of JPEG compression (separate from training) and also apply Instagram filters namely \textit{Aden}, \textit{Brooklyn} and \textit{Clarendon} which we use from an open-source python library - Pilgram~\cite{pilgramfilter}. Some example images from these transformations are shown in Figure~\ref{fig:filters}. 
We report the BRA of different watermarking frameworks after undergoing benign transformations in Table~\ref{tab:fullcompare}.
We find that both StegaStamp and our robust and semi-fragile models can decode secrets with a high BRA for these image transformations. 
We find that \textit{FaceSigns (Robust)}, which does not use malicious transforms during training, is slightly more robust to benign transformations as compared to \textit{FaceSigns (Semi-Fragile)}. However, this improved robustness comes at the cost of being non-fragile to malicious transformations and being able to decode messages with high BRA even for Deepfake manipulations. The model \textit{FaceSigns (No-Transform)} which does not incorporate any benign or malicious transformations during training is fragile to both JPEG compression and malicious transforms as indicated by the low BRA for both methods. 

\begin{figure}[h]
    \centering
    \includegraphics[width=0.9\linewidth]{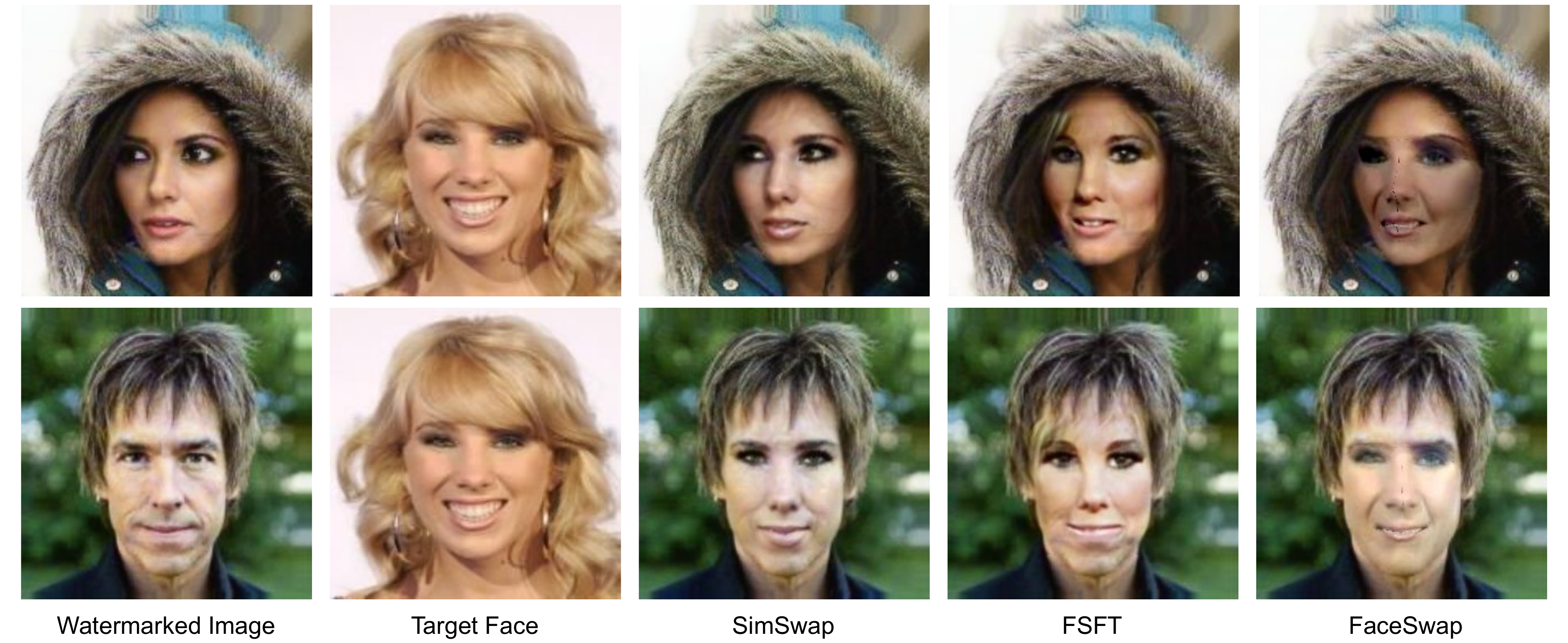}
    \label{fig:deepfakes}
    \caption{Examples of facially manipulated images using SimSwap~\cite{simswap}, FSFT~\cite{fsft} and FaceSwap~\cite{faceswap} techniques used for evaluating the fragility of the watermark.}
\end{figure}
\begin{figure}[h]
    \centering
    \includegraphics[width=1.0\linewidth]{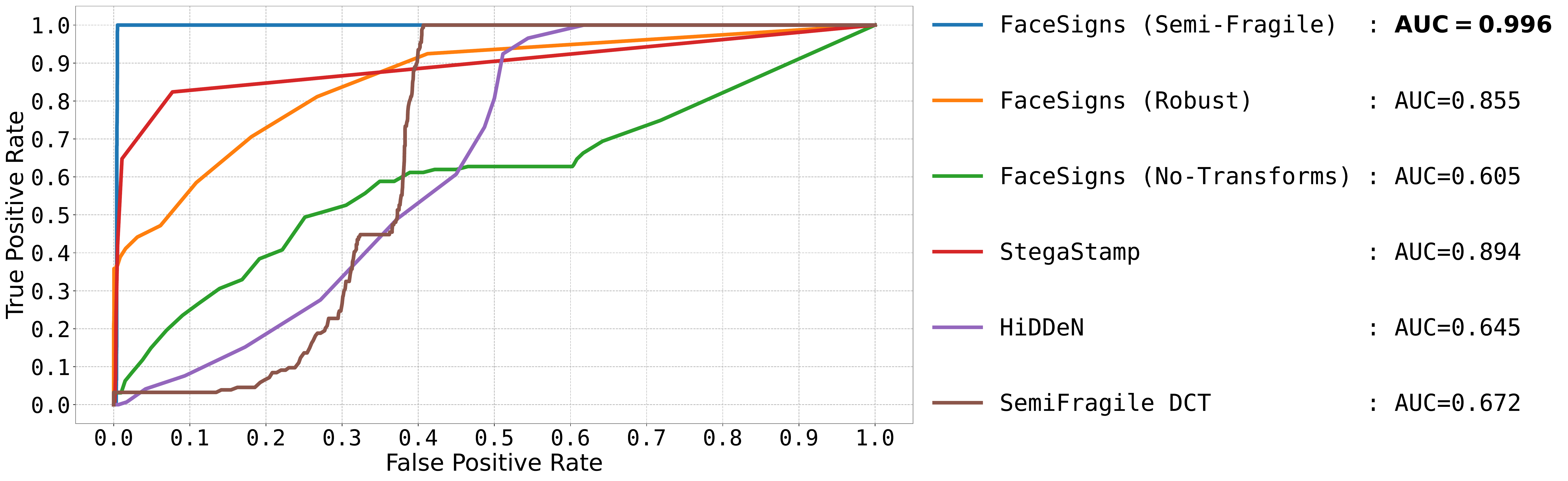}
    \caption{
    Manipulation detection ROC plots and AUC scores for different watermarking techniques. 
    A positive example represents a facially manipulated image while a negative example represents a benign transformed image (all the transformations listed in Table~\ref{tab:fullcompare}). The watermarking framework labels an example as manipulated if the BRA for an image is less than a given threshold.
    }
    \label{fig:roc}
\end{figure}

To evaluate fragility of the watermark against unseen facial manipulations, we apply three face-swapping techniques on the watermarked images: FaceSwap~\cite{faceswap}, SimSwap~\cite{simswap} and Few-Shot Face Translation (FSFT)~\cite{fsft}. FaceSwap~\cite{faceswap} is a computer graphics based technique that swaps the face by aligning the facial landmarks of the two images. SimSwap~\cite{simswap} and FSFT~\cite{fsft} are deep learning based techniques that use CNN encoder-decoder networks trained using adversarial loss to generate Deepfakes.
As reported in 
Table~\ref{tab:fullcompare}, we find that StegaStamp and \textit{FaceSigns (Robust)} can decode signatures from Deepfake images with a high BRA thereby making them unsuitable for authenticating the integrity of digital media. This is understandable since these methods prioritize robustness over fragility. StegaStamp has been shown to be robust to occlusions even though occlusions were not explicitly a part of their set of training transformations. 
As evident by the ROC plots and AUC scores shown in Figure~\ref{fig:roc}, in contrast to prior works, our semi-fragile model  demonstrates robustness to benign transformations while being fragile toward out-of-domain malicious Deepfake transformations, thereby achieving our goal of selective fragility and an AUC score of $0.996$ for manipulation detection. 

\setlength\tabcolsep{2pt}
\begin{table*}[h]
\centering
\begin{tabular}{@{}l|cccccc|ccc@{}}
\multicolumn{1}{r}{} & \multicolumn{6}{c|}{\emph{BRA (\%) - Benign Transforms }} & \multicolumn{3}{c}{\emph{BRA (\%) - Malicious 
Transforms}}
\\

\midrule
Method & None & JPG-75 & JPG-50 & \textit{Aden} & \textit{Brooklyn} & \textit{Clarendon}  & SimSwap~\cite{simswap} & FSFT~\cite{fsft} & FaceSwap~\cite{faceswap} \\
\midrule
SemiFragile DCT~\cite{ho2004semi} & $99.81$ & $56.65$ & $55.04$ & $94.98$ & $96.41$ & $95.06$ & $57.62$ & $57.61$  & $88.59$\\
HiDDeN~\cite{zhu2018hidden} & $97.06$ & $72.71$ & $68.48$ & $94.52$ & $94.52$ & $94.52$ & $85.48$ & $72.33$ & $74.23$ \\
StegaStamp~\cite{tancik2020stegastamp} & $99.92$ & $99.91$ & $99.87$ & $99.84$ & $99.73$ & $99.39$ & $98.34$ & $97.42$ & $97.43$ \\ 
\midrule
FaceSigns (No Transform) & $99.96$ & $50.51$ & $50.07$ & $98.39$ & $99.67$ & $99.65$ & $51.04$ & $52.00$ & $51.36$\\
FaceSigns (Robust) & $ 99.96$ & $99.74$ & $97.26$ & $99.53$ & $99.19$ & $99.37$ & $97.29$ & $89.76$ & $68.99$ \\
\textbf{FaceSigns (Semi-Fragile)} & $99.68$ & $99.49$ & $98.38$ & $97.40$ & $98.34$ & $99.32$ & $64.93$ & $52.21$ & $31.77$ \\
\bottomrule
\multicolumn{1}{c}{}
\end{tabular}
\caption{Bit recovery accuracy (BRA) of different watermarking techniques against benign and malicious transformations. For Benign transformations we consider two JPEG compression levels and three instagram filters --- Aden, Brooklyn and Clarendon. For malicious transforms we consider two GAN based face swapping techniques --- SimSwap~\cite{simswap} and FSFT~\cite{fsft}.
A higher BRA against benign and lower BRA against malicious transforms is desirable to achieve our goal of semi-fragile watermarking.}
\label{tab:fullcompare}
\end{table*}

\subsection{Watermarking images with multiple faces}

\begin{figure}[h]
    \centering
    \includegraphics[width=0.7\linewidth]{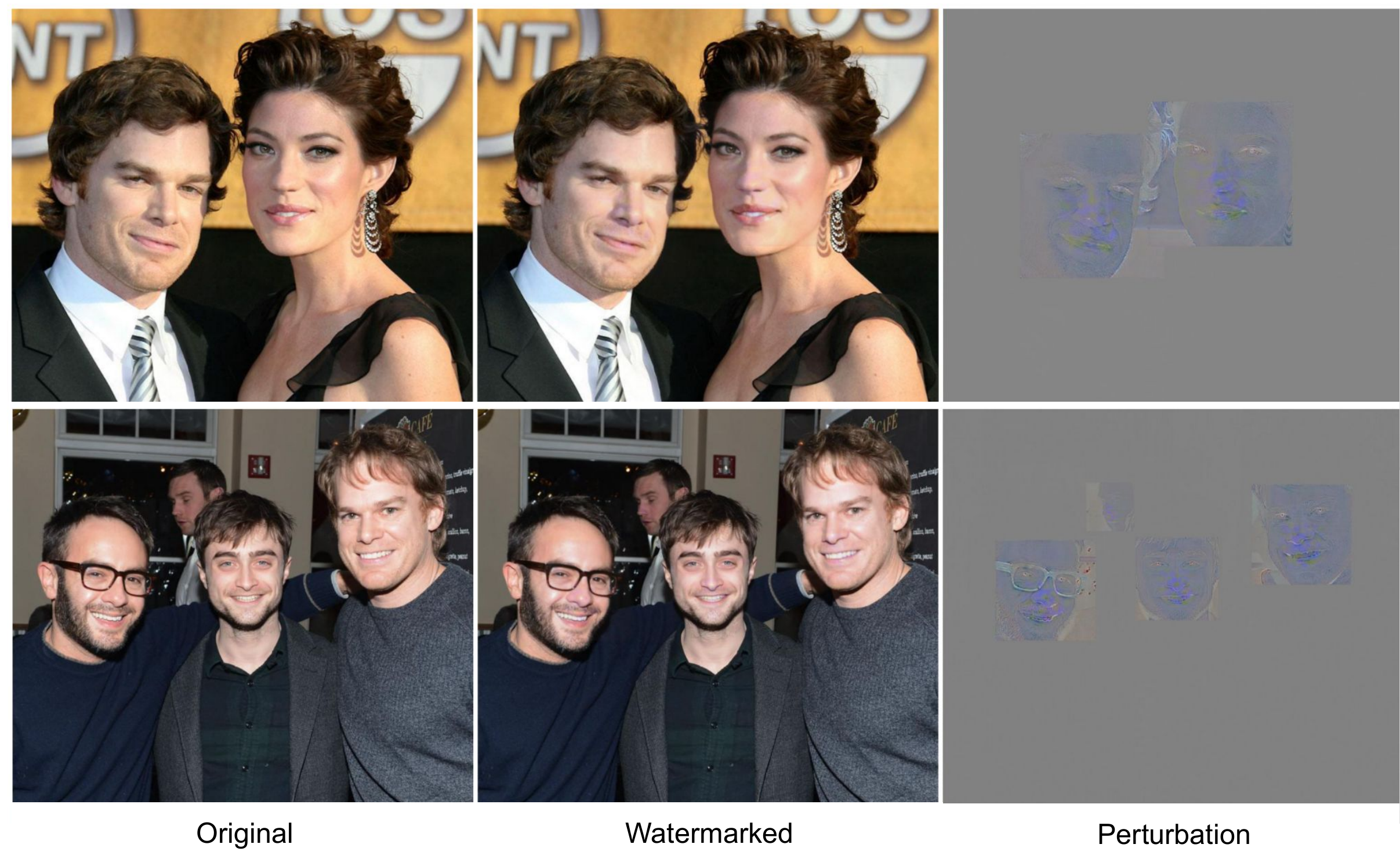}
    
    \caption{Examples of watermarked images with multiple faces using \textit{FaceSigns (Semi-Fragile)}. Our model reliably embeds watermarks into faces present in the foreground and background.
    }
    \label{fig:multiplefaces}
\end{figure}

While our networks are trained on the CelebA dataset that has only one face per image, our watermarking framework can also be applied to images containing multiple faces. To achieve this, we use a face detection model to extract a square bounding box of each face in the image containing multiple faces. The faces from the bounding boxes are cropped out and resized to $256\times256$ and passed as input to our encoder model to embed individual semi-fragile watermarks. The watermarked faces are then resized back to their original size and placed back into the original images. During decoding, a similar process is repeated where the faces are cropped and resized to $256\times256$ before being fed into the decoder. Since the benign transforms during training tolerate small image translations, it makes our watermarking robust to small shifts in the face detection network. 
We conduct experiments on 400 test images containing $2 \text{ to } 6$ faces each from the Celebrity Together dataset~\cite{Zhong18}. Our \textit{FaceSigns (Semi-Fragile)} model achieves a BRA of 99.50\% demonstrating that we can effectively encode and retrieve watermarks embedded in images with multiple faces. Figure~\ref{fig:multiplefaces} shows watermarked images with multiple identities. 

\section{Discussion - Threat Models}
\label{sec:threatmodels}
Both watermark embedding techniques and Deepfake detection systems face adversarial threats from attackers who attempt to bypass the detectors by authenticating manipulated media.
In this section, we discuss some of the threat models faced by our system and how these challenges can be addressed:

\noindent \textbf{\textit{Attack 1.} Querying the decoder network for performing adversarial attacks:} The attacker may query the decoder network with an image to get the decoded message and adversarially perturb the query image until the decoded message matches the target message.\\
\textbf{\textit{Defense:}} 
The attacker does not know what target messages can prove media authenticity since these messages can be kept as a secret and updated frequently. If the attacker gains access to the secret message by querying the decoder with a watermarked image, the encryption key secrecy can prevent the attacker from knowing the target encrypted message for the decoder. Lastly, the decoder network can be hosted securely and can only output a binary label indicating whether the image is authentic or manipulated by matching the decoded secret with the list of trusted secrets.
This would make the decoder's signal unusable for performing adversarial attacks to match a target message out of the total possible $2^{128}$ messages. 

\noindent \textbf{\textit{Attack 2.} Copying the watermark perturbation from one image to another:} The adversary may attempt to extract the added perturbation of the watermark and add it onto a Deepfake image to authenticate the manipulated media. 
\\
\textbf{\textit{Defense:}}
Since FaceSigns generates an image and message specific perturbation, we hypothesize that the same perturbation when applied on alternate images should not be recoverable by the decoder. We verify this hypothesis by conducting an experiment in which we extract added perturbations from 100 watermarked images, and apply extracted perturbation to 100 alternate images. The bit recovery accuracy of such an attack is just 17.6\% which is worse than random prediction.

\noindent \textbf{\textit{Attack 3.} Training a proxy encoder:} The adversary can collect a dataset of original and watermarked images and train a neural network based encoder-decoder image-to-image translation network to map any new image to a watermarked image.\\
\textbf{\textit{Defense:}}
One defense strategy is to only store watermarked images on devices so that an attacker never gains access to pairs of original and watermarked images. 
Also, the above attack can only work if the encoded images all contain the same secret message, so that the adversary can learn a generator for watermarking a new image with the same secret message. To prevent the creation of such a dataset, some bits of the message can be kept dynamic and contain a unique time-stamp and device specific codes so that each embedded bit-string is different.
Regularly updating the trusted message or encryption key is another preventative strategy against such attacks.

\section{Conclusion}
We introduce a deep learning based semi-fragile watermarking system that can certify the integrity of digital images and reliably detect facial tampering. Through our experiments and evaluations, we demonstrate that FaceSigns generates more imperceptible watermarks than previous state-of-the-art methods while upholding the desired semi-fragile characteristics. By carefully designing a fixed set of differentiable benign and malicious transformations during training, our framework achieves generalizability to real-world image transformations and can reliably detect Deepfake facial manipulations unlike prior image watermarking techniques.
FaceSigns can be vital to media authenticators in social media platforms, news agencies and legal offices and help create more trustworthy platforms and establish consumer trust in digital media.




%
%
\bibliographystyle{splncs04}
\bibliography{egbib}
\end{document}